%% file: ms.tex
\documentclass{article}
\usepackage{ijcai18}

\usepackage{times}
\usepackage{soul}
\usepackage[utf8]{inputenc}
\usepackage[small]{caption}

\newcommand{\citeN}[1]{\citeauthor{#1}~(\citeyear{#1})}

\usepackage{array,multirow,hhline}
\usepackage{booktabs}
\usepackage{graphicx}
\usepackage{amsmath,amssymb,amsfonts,dsfont,amsbsy}
\usepackage[svgnames]{xcolor}
\usepackage{url}
\usepackage{multicol}
\usepackage{array}
\usepackage{multirow}
\usepackage{tikz}
\usetikzlibrary{shapes.geometric}
\usetikzlibrary{positioning,shapes,shadows,arrows,calc,mindmap,calc,fit}
\usepackage{amsthm}
\theoremstyle{plain}
\usepackage{microtype}

\newtheorem*{mydef}{Problem Statement}

\title{Neural Networks for Predicting Algorithm Runtime Distributions}
\author{
Katharina Eggensperger, 
Marius Lindauer, 
Frank Hutter
\\ 
University of Freiburg \\
\{eggenspk, lindauer, fh\}@cs.uni-freiburg.de
}

\begin{document}

\maketitle
\input{macros}

\begin{abstract}
Many state-of-the-art algorithms for solving hard combinatorial problems in artificial intelligence (AI) include elements of stochasticity that lead to high variations in runtime, even for a fixed problem instance.
Knowledge about the resulting runtime distributions (RTDs) of algorithms on given problem instances can be exploited in various meta-algorithmic procedures, such as algorithm selection, portfolios, and randomized restarts.
Previous work has shown that machine learning can be used to individually predict mean, median and variance of RTDs.
To establish a new state-of-the-art in predicting RTDs, we demonstrate that the parameters of an RTD should be learned jointly
and that neural networks can do this well by directly optimizing the likelihood of an RTD given runtime observations.
In an empirical study involving five algorithms for SAT solving and AI planning, we show that neural networks predict the true RTDs of unseen instances better than previous methods, and can even do so when only few runtime observations are available per training instance.
\end{abstract}

\section{Introduction}
\label{sec:intro} 

Algorithms for solving hard combinatorial problems often rely on random choices and decisions to improve their performance. For example, randomization helps to escape local optima, enforces stronger exploration and diversifies the search strategy by not only relying on heuristic information. 
In particular, most local search algorithms are randomized~\cite{hoos-sls04} and structured tree-based search algorithms can also substantially benefit from randomization~\cite{gomes-jar00a}. 

The runtimes of randomized algorithms for hard combinatorial problems are well-known to vary substantially, often by orders of magnitude, even when running the same algorithm multiple times on the same instance~\cite{gomes-jar00a,hoos-sls04,hurley-ijcai15a}.
Hence, the central object of interest in the analysis of a randomized algorithm on an instance is its \emph{runtime distribution (RTD)}, in contrast to a single scalar for deterministic algorithms.
Knowing these RTDs is important in many practical applications, such as computing optimal restart strategies~\cite{luby-ipl93}, optimal algorithm portfolios~\cite{gomes-aij01} and the speedups obtained by executing multiple independent runs of randomized algorithms~\cite{hoos-sls04}.

It is trivial to measure an algorithm's empirical RTD on an instance by running it many times to completion, but for new instances
this is of course not practical. Instead, one would like to estimate the RTD for a new instance \emph{without running the algorithm on it}.

There is a rich history in AI that shows that the runtime of algorithms for solving hard combinatorial problems can indeed be predicted to a certain degree~\cite{Bre95,Roberts07learnedmodels,Fink98howto,leyton-brown-acm09a,hutter-aij14a}. These runtime predictions have enabled a wide range of meta-algorithmic procedures, such as algorithm selection~\cite{xu-jair08a}, model-based algorithm configuration~\cite{hutter-lion11a}, generating hard benchmarks~\cite{leyton-brown-acm09a}, gaining insights into instance hardness~\cite{SmiLop12} and algorithm performance~\cite{hutter-lion13a}, and creating cheap-to-evaluate surrogate benchmarks~\cite{eggensperger-ml18a}.

Given a method for predicting RTDs of randomized algorithms, all of these applications could be extended by an additional dimension. Indeed, predictions of RTDs have already enabled applications such as dynamic algorithm portfolios~\cite{gagliolo-amai06a}, adaptive restart strategies~\cite{gagliolo-cp06a,haim-sat09a}, and predictions of the runtime of parallelized algorithms~\cite{arbelaez-ictai16a}.
To advance the underlying foundation of these applications, in this paper we focus on better methods for predicting RTDs. Specifically, our contributions are as follows:
\begin{enumerate}
  \item We compare different ways of predicting RTDs and demonstrate that neural networks (NNs) can jointly predict all parameters of various parametric RTDs, yielding RTD predictions that are superior to those of previous approaches (which predict the RTD's parameters independently).
  \item We propose \emph{DistNet}, a practical NN for predicting RTDs, and discuss the bells and whistles that make it work.
  \item We show that DistNet achieves substantially better performance than previous methods when trained only on a few observations per training instance.
\end{enumerate}

\section{Related Work}
\label{sec:relwor}

The rich history in predicting algorithm runtimes focuses on predicting mean runtimes, with only a few exceptions.
\citeN{hutter-cp06a} predicted the single distribution parameter of an exponential RTD and \citeN{arbelaez-ictai16a} predicted the two parameters of log-normal and shifted exponential RTDs with independent models. In contrast, we \emph{jointly} predict multiple RTD parameters (and also show that the resulting predictions are better than those by independent models). 

The work most closely related to ours is by \citeN{gagliolo-icann05a}, who proposed to use NNs to learn a distribution of the time left until an algorithm solves a problem based on features describing the algorithm's current state and the problem to be solved; they used these predictions to dynamically assign time slots to algorithms. In contrast, we use NNs to predict RTDs for unseen problem instances.

All existing methods for predicting runtime on unseen instances base their predictions on \emph{instance features} that numerically characterize problem instances. In particular in the context of algorithm selection, these instance features have been proposed for many domains of hard combinatorial problems, such as propositional satisfiability~\cite{nudelman-cp04} and AI planning~\cite{fawcett-icaps14a}.
To avoid this manual step of feature construction, \citeN{loreggia-aaai16} proposed to directly use the text format of an instance as the input to a NN to obtain a numerical representation of the instance. Since this approach performed a bit worse than manually constructed features, in this work we use traditional features, but in principle our framework works with any type of features.

\section{Problem Setup}
\label{sec:pipeline}

The problem we address in this work can be formally described as follows:

\begin{mydef}[Predicting RTDs]
Given 
\begin{itemize}
  \item a randomized algorithm $A$
  \item a set of instances $\insts_{train} = \{\inst_1, \dots, \inst_n\}$
  \item for each instance $\inst \in \insts_{train}$:
  \begin{itemize}
    \item $m$ instance features $\feat(\inst) = [\mathrm{f}(\inst)_1, \dots, \mathrm{f}(\inst)_m]$
    \item runtime observations $\vec{t}(\inst) = \langle t(\inst)_1, \dots, t(\inst)_{k}\rangle$ obtained by executing $A$ on $\inst$ with $k$ different seeds,
  \end{itemize}  
\end{itemize} the goal is to learn a model that can predict $A$'s RTD well for unseen instances $\inst_{n+1}$ with given features $\feat(\inst_{n+1})$. 
\end{mydef}

Following the typical approach in the literature~\cite{hutter-cp06a,arbelaez-ictai16a}, we address this problem in two steps:
\begin{enumerate}
  \item Determine a parametric family $\mathcal{D}$ of RTDs with parameters $\dparam$ that fits well across training instances;
  \item Fit a machine learning model that, given a new instance and its features, predicts $\mathcal{D}$'s parameters $\dparam$ on that instance.
\end{enumerate} 

\begin{figure*}[t]
\begin{minipage}[t]{0.7\textwidth}
\centering
\input{pipeline}
\caption{Our pipeline for predicting RTDs. Upper part: training; lower part: test.
\label{fig:pipeline}}
\end{minipage}
\begin{minipage}[t]{0.25\textwidth}
\centering
\includegraphics[width=1\textwidth]{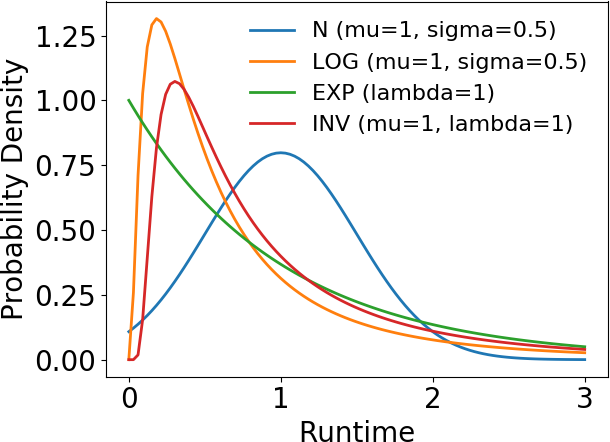}
\caption{Different RTD families.}
\label{fig:art_rtd}
\end{minipage}
\end{figure*}
Figure~\ref{fig:pipeline} illustrates the pipeline we use for training these RTD predictors and using them on new instances.

\subsection{Parametric Families of RTDs}
\label{sec:families}

\begin{table}[t]
\centering
\footnotesize
\setlength{\tabcolsep}{1.3mm}
\begin{tabular}{llc}
\toprule
Distribution & Param. & PDF\\
\midrule
Normal (N) & $\mu, \sigma$ & $\frac{1}{\sqrt{2\pi\sigma^2}} e^{ -\frac{(x-\mu)^2}{2\sigma^2} }$\\
Lognormal (LOG) & $s, \sigma$ & $\frac{1}{\sigma x \sqrt{2\pi}} e^{ -\frac{1}{2} \left(\frac{\log(x) - \log(s)}{\sigma}\right)^2}$\\
Exponential (EXP) & $\beta$ & $\frac{1}{\beta} e^{\left(-\frac{x}{\beta}\right)}$\\
Inverse Normal (INV) & $\mu, \lambda$ & $\left(\frac{\lambda}{2 \pi x^3}\right)^{\frac{1}{2}} e^{\left(\frac{-\lambda(x-\mu)^2}{2 x \mu^2}\right)}$\\
\bottomrule
\end{tabular}
\caption{Considered RTD families}
\label{tab:families}
\end{table}

We considered a set of $4$ parametric probability distributions (shown in Table~\ref{tab:families} with exemplary instantiations shown in Figure~\ref{fig:art_rtd}), most of which have been widely studied to describe the RTDs of combinatorial problem solvers~\cite{frost-aaai97a,gagliolo-cp06a,hutter-cp06a}. 

First, we considered the Normal distribution (N) as a baseline, due to its widespread use throughout the sciences.

Since the runtimes of hard combinatorial solvers often vary on an exponential scale (likely due to the $\mathcal{NP}$-hardness of the problems studied), a much better fit of empirical RTDs is typically achieved by a lognormal distribution (LOG); this distribution is attained if the logarithm of the runtimes is normal-distributed and has been shown to fit empirical RTDs well in previous work~\cite{frost-aaai97a}. 

Another popular parametric family from the literature on RTDs is the exponential distribution (EXP), which tends to describe the RTDs of many well-behaved stochastic local search algorithms well~\cite{hoos-sls04}. It is the unique family with the property that the probability of finding a solution in the next time interval (conditional on not having found one yet) remains constant over time. 

By empirically studying a variety of alternative parametric families,
we also found that an inverse Normal distribution (INV) tends to fit RTDs very well.
By setting its $\lambda$ parameter close to infinity, it can also be made to resemble a normal distribution. Like LOG and EXP, this flexible distribution can model the relatively long tails of typical RTDs of randomized combinatorial problem solvers quite well. 

\subsection{Quantifying the Quality of RTDs}
\label{sec:quantqal}

To measure how well a parametric distribution $\mathcal{D}$ with parameters $\dparam$ fits our empirical runtime observations $\vec{t}(\inst) = \langle{}t(\inst)_1, \dots, t(\inst)_k\rangle{}$ (the \emph{empirical RTD}),
we use the likelihood $\mathcal{L_{\mathcal{D}}}$ of parameters $\dparam$ given all observations $\vec{t}(\inst)$, which is equal to the probability of the observations under distribution $\mathcal{D}$ with parameters $\dparam$:
\begin{equation}
\mathcal{L_{\mathcal{D}}}(\dparam \mid t(\inst)_1, \dots, t(\inst)_k) = \prod^k_{i=1} p_{\mathcal{D}}(t(\inst)_i \mid \dparam).
\end{equation}
Consequently, when estimating the parameters of a given empirical RTD, we use a maximum-likelihood fit.
For numerical reasons, as is common in machine learning, we use the negative log-likelihood (\nllh{}) as a loss function to be minimized:
\begin{equation}
\label{eqn:loss}
\!-\!\log\!\mathcal{L_{\mathcal{D}}}(\dparam \!\mid\! t(\inst)_1, \dots, t(\inst)_k) \!= \!- \sum^k_{i=1} \log p_{\mathcal{D}}(t(\inst)_i | \dparam).\!
\end{equation}
Since each instance $\inst \in \insts$ results in an RTD, we measure the quality of a parametric family of RTDs for a given instance set by averaging over the \nllh{}s of all instances.

One problem of Eq.~(\ref{eqn:loss}) is that it weights easy instances more heavily: if two RTDs have the same shape but differ in scale by a factor of 10 due to one instance being 10 times harder, the PDF for the easier instance is 10 times larger (to still integrate to 1). 
To account for that, for each instance, we multiply the likelihoods with the maximal observed runtime, and use the resulting metric to select a distribution family for a dataset at hand and to compare the performance of our models:

\small
\begin{eqnarray}
\frac{1}{|\insts|} \sum_{\inst \in \insts} - \log \left( \mathcal{L}_{\mathcal{D}}(\dparam \mid t(\inst)_1, \ldots, t(\inst)_k) \cdot \max_{i\in \{1\ldots k\}} t(\inst)_i \right)  \\
= - \frac{1}{|\insts|} \sum_{\inst \in \insts} \left( \left( \sum^k_{i=1} \log p_{\mathcal{D}}(t(\inst)_i | \dparam) \right) + \log \max_{i\in \{1\ldots k\}} t(\inst)_i\right).
\label{eq:normnllh}
\end{eqnarray}
\normalsize

\section{Joint Prediction of multiple RTD Parameters}
\label{sec:predict}

Having selected a parametric family of distributions, the last part of our pipeline is to fit an RTD predictor for new instances as formally defined in Section~\ref{sec:pipeline}.
In the following, we briefly discuss how traditional regression models have been used for this problem, and why this optimizes the wrong loss function. We then show how to obtain better predictions with NNs and introduce DistNet for this task.   

\subsection{Generalizing from Training RTDs}
A straightforward approach for predicting parametric RTDs based on standard regression models is to fit the RTD's parameters $\dparam(\inst)$ for each training instance $\inst$, and to then train a regression model on data points $\langle{}\feat(\inst),\dparam(\inst)\rangle{}_{\inst\in\insts_{\text{train}}}$ that directly maps from instance features to RTD parameters. There are two variants to extend these approaches to the problem of predicting multiple parameters of RTDs governed by $p>1$ parameters (e.g. $s$ and $\sigma$ for LOG): (1)~fitting $p$ independent regression models, or (2)~fitting a multi-output model with $p$ outputs.
These approaches have been used before based on Gaussian processes~\cite{hutter-cp06a}, linear regression~\cite{arbelaez-ictai16a} and random forests~\cite{hutter-aij14a,hurley-ijcai15a}

However, we note that these variants measure loss in the space of the distribution parameters $\dparam$ as opposed to the true loss in Equation~(\ref{eqn:loss})
and that both variants require fitting RTDs on each training instance, making the approach inapplicable if we, e.g., only have access to a few runs for each of a thousands of instances.
Now, we show how NNs can be used to solve both of these problems.

\subsection{Predictions with Neural Networks}
\label{ssec:DistNetDesc}

NNs have recently been shown to achieve state-of-the-art performance 
for many supervised machine learning problems as large data sets became available, 
e.g., in image classification and segmentation, speech processing and natural language processing. 
For a thorough introduction, we refer the interested reader to~\citeN{goodfellow-mit16a}.
Here, we apply NNs to RTD prediction.

\paragraph{Background on Neural Networks.}
NNs can approximate arbitrary functions by defining a mapping \mbox{$y = f(x; W)$} where $W$ are the weights to be learnt during training to approximate the function.
In this work we use a fully-connected feedforward network, which can be described as an acyclic graph that connects nonlinear transformations $g$ in a chain, from layer to layer. For example, a NN with two hidden layers that predicts $y$ for some input $x$ can be written as:\footnote{We ignore bias terms for simplicity of exposition.}
\begin{equation}
y = g^{out}\left(g^{(2)}\left(g^{(1)}\left(x W^{(1)}\right)W^{(2)}\right)W^{(3)}\right),
\end{equation}
with $W^{(j)}$ denoting trainable network weights and $g^{(j)}$ (the so-called activation function) being a nonlinear transformation applied to the weighted outputs of the $j$-th layer. We use the $\exp(\cdot)$ activation function for $g^{(out)}$ to constrain all outputs to be positive.

NNs are usually trained with stochastic gradient descent (SGD) methods using backpropagation to effectively obtain gradients of a task-specific loss function for each weight.

\paragraph{Neural Networks for predicting RTDs.}
We have one input neuron for each instance feature,
and we have one output neuron for each distribution parameter.
To this end, we assume that we know the best-fitting distribution family
from the previous step of our pipeline.

We train our networks to directly minimize the \nllh{} of the predicted distribution parameters given our observed runtimes.
Formally, for a given set of observed runtimes and instance features, we minimize the following loss function in an end-to-end fashion:
\begin{equation}
J(W) \propto -\sum_{\inst\in\Pi_{\text{train}}} \sum_{i=1}^{k} log{\mathcal{L}_{\mathcal{D}}\left(\dparam_{W} | \feat(\inst), t(\inst)_i \right)}.
\end{equation}
Here, $\dparam_{W}$ denotes the values of the distribution parameters obtained in the output layer given an instantiation $W$ of the NN's weights.
This optimization process, which targets exactly our loss function of interest (Eq.~(\ref{eqn:loss})), allows to effectively predict all $p$ distribution parameters jointly. 
Since predicted combinations are judged directly by their resulting \nllh{}, the optimization process is driven to find \emph{combinations that work well together}.
This end-to-end optimization process is also more general as it removes the need of fitting an RTD on each training instance and thereby enables using an arbitrary set of algorithm performance data for fitting the model.  

\paragraph{DistNet: RTD predictions with NNs in practice.}
Unfortunately, training an accurate NN in practice can be tricky and requires manual attention to many details, including the network architecture, training procedure, and other hyperparameter settings.

Specifically, to preprocess our runtime data $\langle \feat(\inst_i), t(\inst_i)_{\{1\ldots k\}} \rangle_{i \in 1 \ldots n}$, 
we performed the following steps:

\begin{enumerate}
  \item We removed all (close to) constant features.
  \item For each instance feature type, we imputed missing 
  values (caused by limitations during feature computation) by the median of the known instance features.
  \item We normalized each instance feature to mean $0$ and standard deviation $1$.
  \item We scaled the observed runtimes in a range of $[0,1]$ by dividing it by the maximal observed runtime across all instances. 
This also helps the NN training to converge faster.
\end{enumerate}

\noindent For training DistNet, we considered the following aspects:

\begin{enumerate}
  \item Our first networks tended to overfit the training data if the training data set was too small and the network too large. Therefore we chose a fairly small NN with $2$ hidden layers each with $16$ neurons.   
  In preliminary experiments we found that larger networks tend to achieve slightly better performance on our largest datasets, but we decided to strive for simplicity.
  \item We considered each runtime observation as an individual data sample.
  \item We shuffled the runtime observations (as opposed to, e.g., using only data points from a single instance in each batch) and used a fairly small batch size of $16$ to reduce the correlation of the training data points in each batch.
  \item Our loss function can have very large gradients because slightly suboptimal RTD parameters can lead to likelihoods close to zero (or a very large \nllh{}). Therefore, we used a fairly small initial learning rate of $1e^{-3}$ exponentially decaying to $1e^{-5}$ and used gradient clipping~\cite{pascanu-iclm13a} on top of it.
\end{enumerate}
\noindent Besides that, we used common architectural and parameterization choices: \textit{tanh} as an activation function, SGD for training, batch normalization, and a $\mathbf{L}2$-regularization of $1e^{-4}$. We call the resulting neural network \emph{DistNet}.
\section{Experiments}
\label{sec:exp}

In our experiments, we study the following research questions:

\begin{description}
\item[Q1] Which of the parametric RTD families we considered best describe the empirical RTDs of the SAT and AI planners we study?
\item[Q2] How do DistNet's joint predictions of RTD parameters compare to those of popular random forest models?
\item[Q3] Can DistNet learn to predict entire RTDs based on training data that only contains a few observed runtimes for each training instance?
\end{description}

\subsection{Experimental Setup}

\begin{table}[t]
\centering
\footnotesize
\begin{tabular}{lrrr}
\toprule
Scenario & \#instances & \#features & cutoff [sec] \\
\midrule
\claspfact{}$^2$     & 2000  & 102 & 5000 \\
\cvvar{}$^2$         & 10011 &  46 & 60  \\
\spearhard{}$^2$     & 8076  &  91 & 5000 \\
\yalsathard{}$^2$    & 11747 &  91 & 5000 \\
\spearsmall{}$^2$    & 11182 &  76 & 5000 \\
\yalsatsmall{}$^2$   & 11182 &  76 & 5000 \\
\zeno{}$^3$          & 3999  & 165 & 300 \\
\bottomrule
\end{tabular}
\caption{Characteristics of the used data sets. \label{tab:data}
}
\end{table}
%
\addtocounter{footnote}{1}
\footnotetext{Run on a compute cluster with nodes equipped with two Intel Xeon E5-2630v4 and $128$ GB memory running CentOS 7.}

\addtocounter{footnote}{1}
\footnotetext{Run on a compute cluster with nodes equipped with two Intel Xeon E5-2650v2 and $64$ GB memory running Ubuntu 14.04.}

We focus on predicting the RTDs of $5$ well-studied algorithms, each evaluated on a different set of problem instances from two different domains:
\begin{description}
	\item[\claspfact{}] is based on the tree-based CDCL solver \clasp~\cite{gebser-ai12} which we ran on SAT-encoded factorization problems instances.
	\item[\cvvar{}] is based on the local search SAT solver \saps{}~\cite{saps}. The SAT instances are randomly generated with a varying clause-variable ratio.
	\item[\spearsmall{}/\yalsatsmall{}] are based on the tree-search SAT solver \spear{}~\cite{spear} and on the local search SAT solver \yalsat{}~\cite{yalsat}, a combination of different variants of \probsat{}~\cite{probSAT}. The instances are SAT-encoded small world graph coloring problems~\cite{gent-aaai99}.
    \item[\spearhard{}/\yalsathard{}] are based on the same solvers as \spearsmall{} and \yalsatsmall{}. The SAT instances encode quasigroup completion instances~\cite{gomes-aaai97}.
    \item[\zeno{}] is based on the local search AI-planning solver \lpg{}~\cite{gerevini-aips02}. The instances are from the \emph{zenotravel} planning domain~\cite{penberthy-aaai94}, which arise in a version of route planning.
\end{description}
To gather training data, we ran each algorithm (with default parameters) with $100$ different seeds on each instance\footnote{We removed instances for which no instance features could be computed and only considered instances which could always be solved within a cutoff limit.}. This resulted in the 7 datasets shown in Table~\ref{tab:data}. 
We used the open-source neural network library~\emph{keras}~\cite{chollet-15a} for our NN, \emph{scikit-learn}~\cite{scikit-learn} for the RF implementation, and \emph{scipy}~\cite{jones-01a} for fitting the distributions.~\footnote{Code and data can be obtained from here: \\ \url{http://www.ml4aad.org/distnet}}

\subsection{Q1: Best RTD Families}

\begin{figure}[tbp]
\centering
\begin{tabular}{@{\hskip 0.1mm}c@{\hskip 0.1mm}c@{\hskip 0.1mm}c@{\hskip 0.1mm}c@{\hskip 0.1mm}}
\small{{\claspfact{}}} & \small{{\cvvar{}}} & \small{{\zeno{}}} & \\
\includegraphics[width=0.12\textwidth]{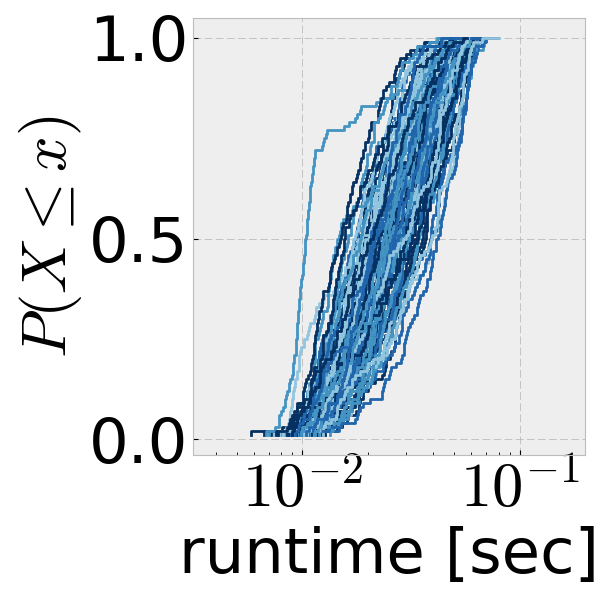} &
\includegraphics[width=0.12\textwidth]{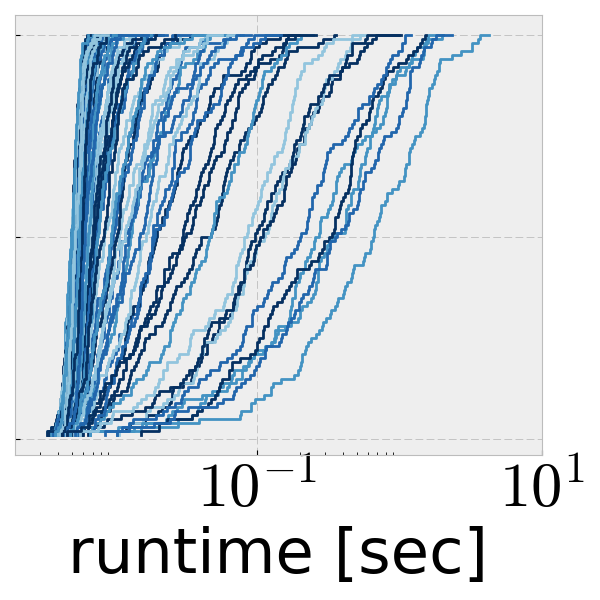} & 
\includegraphics[width=0.12\textwidth]{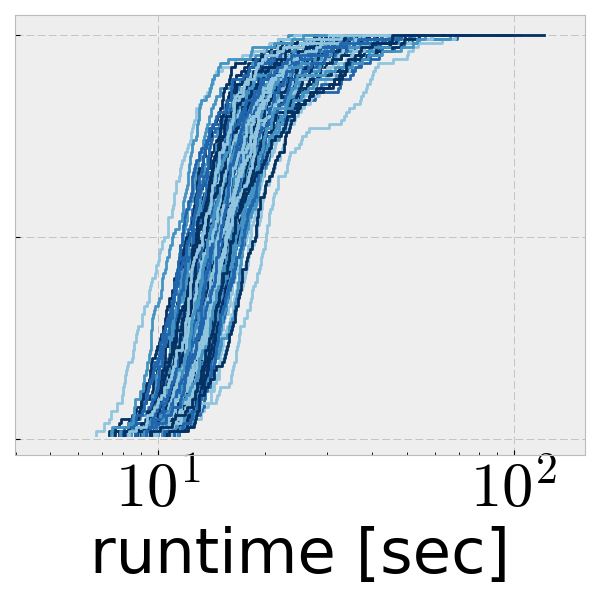} & \\
\midrule
\small{{\spearhard{}}} & \small{{\yalsathard{}}} & \small{{\spearsmall{}}} & \small{{\yalsatsmall{}}} \\
\includegraphics[width=0.12\textwidth]{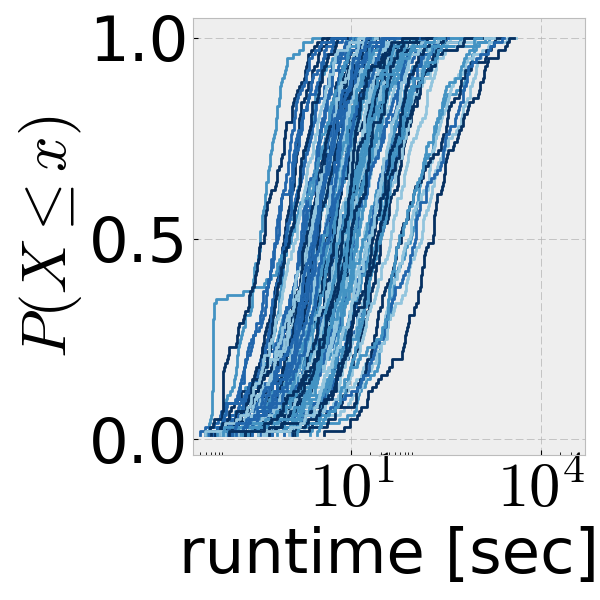} &
\includegraphics[width=0.12\textwidth]{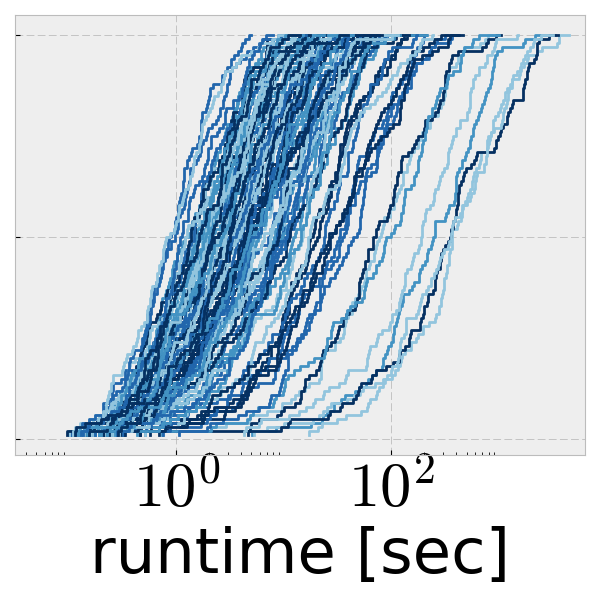} &
\includegraphics[width=0.12\textwidth]{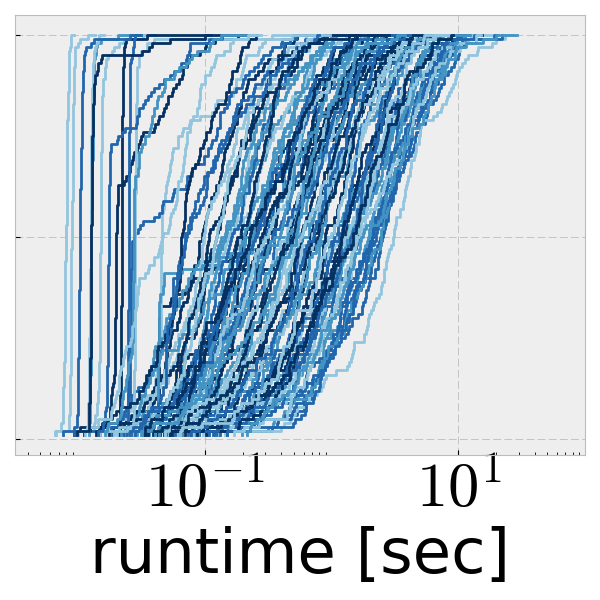} &
\includegraphics[width=0.12\textwidth]{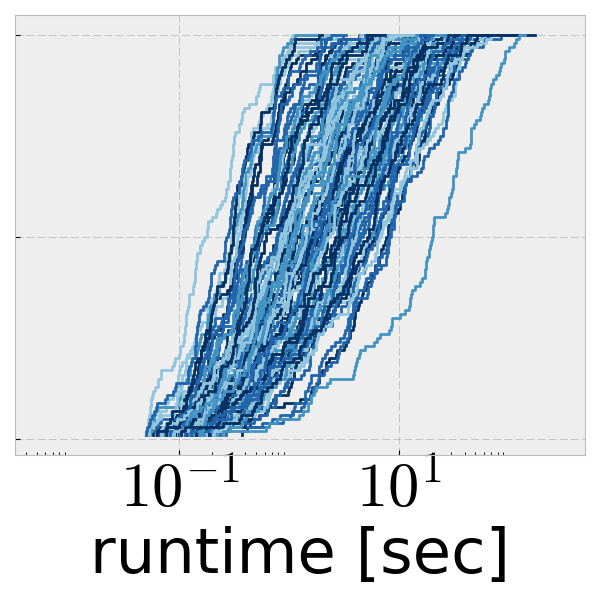} \\
\end{tabular}
\caption{A subset of empirical CDFs observed when running the default configuration of an algorithm 100 times.
\label{fig:picto_cdf}}
\end{figure}

Figure~\ref{fig:picto_cdf} shows some exemplary CDFs of our empirical RTDs; each line is the RTD on one of the instances.
The different algorithms' RTDs show different characteristics with most instances having a long right tail and a short left tail. The RTDs of \claspfact{} in contrast have a short left and right tail. Also, for some instances from \cvvar{} and \spearsmall{} the runtimes were very short and similar, causing almost vertical CDFs.
 
Table~\ref{tab:nllhkstest} shows a quantitative evaluation of the different RTD families we considered. Next to the normalized \nllh{} (see Equation~\eqref{eq:normnllh}), we followed~\citeN{arbelaez-ictai16a} and evaluated the Kolmogorov-Smirnov (KS) test as a goodness of fit statistical test. 
The KS-statistic is based on the maximal distance between an empirical distribution and the cumulative distribution function of a reference distribution. 
To aggregate the test results across instances, we count how often the KS-test rejected the null-hypothesis that our measured $t(\inst)_i$ are drawn from a reference RTD. 

\begin{table}[tbp]
\small
\centering
\setlength{\tabcolsep}{1.0mm}
\begin{tabular}{@{\hskip 0.1mm}l@{\hskip 1mm}
                c@{\hskip 1.2mm}c@{\hskip 1.2mm}c@{\hskip 1.2mm}c | @{\hskip 1mm}
                c@{\hskip 1.2mm}c@{\hskip 1.2mm}c@{\hskip 1.0mm}c@{\hskip 1.2mm}}

\toprule
\multicolumn{1}{c}{} & \multicolumn{4}{c}{$-\!\log\!\mathcal{L_{\mathcal{D}}}(\dparam \!\mid\! t(\inst))$} & \multicolumn{4}{c}{KS: (\%$p) \leq 0.01$} \\
\hhline{-|----|----} \\[-3.4mm]
 \multirow{2}{*}{\small{\claspfact{}}}   & INV    & LOG    & N      & EXP    & INV  & LOG  & N    & EXP   \\
                                         & -0.35  & -0.35  & -0.29  & 0.29   & 12.0 & 10.2 & 15.0 & 100 \\[-0.5mm]
\hhline{-|----|----} \\[-3.4mm]
\multirow{2}{*}{\small{\cvvar{}}}        & LOG    & INV    & N      & EXP    & LOG  & INV  & N    & EXP   \\
                                         & -0.88  & -0.88  & -0.75  & 0.26  & 0.1  & 4.0  & 20.1 & 87.5  \\[-0.5mm]
\hhline{-|----|----} \\[-3.4mm]
 \multirow{2}{*}{\small{\spearhard{}}}   & LOG    & EXP    & INV    & N      & LOG  & EXP  & INV  & N     \\
                                         & -1.20  & -1.14  & -1.10  & -0.41  & 1.1  & 22.6 & 52.1 & 99.3  \\[-0.5mm]
\hhline{-|----|----} \\[-3.4mm]
 \multirow{2}{*}{\small{\yalsathard{}}}  & LOG    & INV    & EXP    & N      & LOG  & INV  & EXP  & N     \\
                                         & -0.78  & -0.78  & -0.66  & -0.32  & 0.0  & 6.8  & 46.5 & 80.1  \\[-0.5mm]
\hhline{-|----|----} \\[-3.4mm]
 \multirow{2}{*}{\small{\spearsmall{}}}  & LOG    & INV    & EXP    & N      & LOG  & INV  & EXP  & N    \\
                                         & -0.93  & -0.90  & -0.71  & -0.41  & 15.2 & 26.7 & 24.5 & 79.0 \\[-0.5mm]
\hhline{-|----|----} \\[-3.4mm]
 \multirow{2}{*}{\small{\yalsatsmall{}}} & LOG    & INV    & EXP    & N      & LOG  & INV  & EXP  & N    \\
                                         & -0.94  & -0.91  & -0.89  & -0.30  & 0.0  & 25.3 & 14.0 & 98.0 \\[-0.5mm]
\hhline{-|----|----} \\[-3.4mm]
\multirow{2}{*}{\small{\zeno{}}}         & LOG    & INV    & N      & EXP    & LOG  & INV  & N    & EXP   \\
                                         & -0.90  & -0.90  & -0.62  & -0.08  & 12.7 & 20.2 & 79.2 & 100 \\[-0.5mm]
\bottomrule
\end{tabular}
\caption{Results for fitted distributions: average \nllh{} across instances and percentage of rejected distributions according to a KS-test ($\alpha=0.01$ without multiple testing correction). For each scenario, we report result for all distributions ranked by the \nllh{}.
For both metrics, smaller numbers are better.
\label{tab:nllhkstest}}
\end{table}

Overall, the best fitted distributions closely resembled the true empirical RTDs, with a rejection rate of the KS-test of at most 15.2\%.
Hence, on most instances the best fitted distributions were not statistically significantly different from the empirical ones. 
For most scenarios the log-normal distribution (LOG) performed best, closely followed by the inverse Normal (INV). The Normal (N) and exponential (EXP) distributions performed worse for all scenarios.
On \spearsmall{}, the KS-test showed the most statistically significant differences for the best fitting distribution since the RTDs for some instances only have a small variance (see Figure~\ref{fig:picto_cdf}) and cannot be perfectly approximated by the distributions we considered. Still, these distributions achieved good \nllh{} values.

\subsection{Q2: Predicting RTDs}

Next, we turn to the empirical evaluation of our DistNet and compare it to previous approaches. Since random forests (RFs) have been shown to perform very well for standard runtime prediction tasks~\cite{hutter-aij14a,hurley-ijcai15a}, we experimented with them in two variants: fitting a multi-output RF (\mrf{}) and fitting multiple independent RFs, one for each distribution parameter (\irf{}).
We trained DistNet as described in Section~\ref{ssec:DistNetDesc} and limit the training to take at most 1h or 1000 epochs, whichever was less. As a gold standard, we report the \nllh{} obtained by a maximum likelihood fit to the empirical RTD ("fitted" in Table~\ref{tab:nllhkstest}).

\begin{table}[tb]
\centering
\footnotesize
\setlength{\tabcolsep}{1.5mm}
\begin{tabular}{lccr|rrc}
\toprule
Scenario                          & dist                 & {}           & fitted  & \irf{} & \mrf{} & DistNet \\
\midrule
\multirow{4}{*}{\claspfact{}}     & \multirow{2}{*}{INV} & \scriptsize{train} & -0.35 & -0.26 & \textbf{-0.28}  &         -0.24  \\
                                  &                      & \scriptsize{test}  & -0.35 & -0.04 &         -0.09   & \textbf{-0.16} \\
\hhline{~---|---} \\[-3.4mm]
\multirow{2}{*}{}                 & \multirow{2}{*}{LOG} & \scriptsize{train} & -0.35 & -0.30 & \textbf{-0.30} &         -0.24  \\                                  
                                  &                      & \scriptsize{test}  & -0.35 & -0.14 &         -0.13  & \textbf{-0.14} \\
\midrule
\multirow{4}{*}{\small{\cvvar{}}} & \multirow{2}{*}{LOG} & \scriptsize{train} & -0.88  & 0.66 & \textbf{-0.68} &          -0.54  \\
                                  &                      & \scriptsize{test}  & -0.88  & 0.99 &         -0.29  & \textbf{-0.52} \\
\hhline{~---|---} \\[-3.4mm]
\multirow{2}{*}{}                 & \multirow{2}{*}{INV} & \scriptsize{train} & -0.88 & -0.46 & \textbf{-0.57} &         -0.54  \\
                                  &                      & \scriptsize{test}  & -0.88 &  0.22 &         -0.09  & \textbf{-0.54} \\
\midrule
\multirow{4}{*}{\small{\spearhard{}}} & \multirow{2}{*}{LOG} & \scriptsize{train} & -1.20 & -1.09 & \textbf{-1.13} & -1.11 \\
                                      &                      & \scriptsize{test}  & -1.20 & -1.00 & -0.96 & \textbf{-1.10} \\
\hhline{~---|---} \\[-3.4mm]
\multirow{2}{*}{}                     & \multirow{2}{*}{EXP} & \scriptsize{train} & -1.14 & \textbf{-1.05} & \textbf{-1.05} &         -0.93  \\
                                      &                      & \scriptsize{test}  & -1.14 &         -0.88  &         -0.88  & \textbf{-0.91} \\
\midrule
\multirow{4}{*}{\small{\yalsathard{}}}& \multirow{2}{*}{LOG} & \scriptsize{train} & -0.78 & -0.50 & \textbf{-0.77} &         -0.76  \\
                                      &                      & \scriptsize{test}  & -0.78 & -0.49 &         -0.74  & \textbf{-0.75} \\
\hhline{~---|---} \\[-3.4mm]
\multirow{2}{*}{}                     & \multirow{2}{*}{INV} & \scriptsize{train} & -0.78 &  -0.68 & \textbf{-0.77} &         -0.74  \\
                                      &                      & \scriptsize{test}  & -0.78 &  -0.66 &         -0.73  & \textbf{-0.74} \\
\midrule
\multirow{4}{*}{\small{\spearsmall{}}}& \multirow{2}{*}{LOG} & \scriptsize{train} & -0.93 & 2.46 & -0.23 & \textbf{-0.48} \\
                                      &                      & \scriptsize{test}  & -0.93 & 0.82 &  0.26 & \textbf{-0.47} \\                       
\hhline{~---|---} \\[-3.4mm]
\multirow{2}{*}{}                     & \multirow{2}{*}{INV} & \scriptsize{train} & -0.90 & 3.60 & 3.32 & \textbf{-0.33} \\
                                      &                      & \scriptsize{test}  & -0.90 & 3.27 & 2.58 & \textbf{-0.32} \\
\midrule
\multirow{4}{*}{\small{\yalsatsmall{}}}& \multirow{2}{*}{LOG}& \scriptsize{train} & -0.94 & -0.81 & \textbf{-0.88} &         -0.81  \\
                                      &                      & \scriptsize{test}  & -0.94 & -0.69 &         -0.71  & \textbf{-0.81} \\
\hhline{~---|---} \\[-3.4mm]
\multirow{2}{*}{}                     & \multirow{2}{*}{INV} & \scriptsize{train} & -0.91 & -0.80 & \textbf{-0.86} &         -0.76  \\
                                      &                      & \scriptsize{test}  & -0.91 & -0.68 &         -0.69  & \textbf{-0.76} \\
\midrule
\multirow{4}{*}{\zeno{}}          & \multirow{2}{*}{LOG} & \scriptsize{train} & -0.90 & -0.89 & \textbf{-0.89} &         -0.85  \\
                                  &                      & \scriptsize{test}  & -0.90 & -0.85 &         -0.84  & \textbf{-0.85} \\
\hhline{~---|---} \\[-3.4mm]
\multirow{2}{*}{}                 & \multirow{2}{*}{INV} & \scriptsize{train} & -0.90 & -0.84 & \textbf{-0.87} &          -0.84  \\
                                  &                      & \scriptsize{test}  & -0.90 & -0.72 &         -0.80  &  \textbf{-0.84} \\ 
\bottomrule
\end{tabular}
\caption{Averaged \nllh{} achieved for predicting RTDs for unseen instances. We report the average across a 10-fold cross-validation with the first line for each dataset being the performance on the training data and the second line being the performance on the test data. For each dataset, we picked the two best-fitting RTD families (according to \nllh{}; see Table~\ref{tab:nllhkstest}) and highlight the best predictions.
\label{tab:nllh}}
\end{table}

Table~\ref{tab:nllh} shows the \nllh{} achieved using a $10$-fold cross-validation, i.e., we split the instances into $10$ disjoint sets, train our models on all but one subset and measure the test performance on the left out subset. We report the average performance (see Eq.\eqref{eq:normnllh}) on train and test data across all splits for the two best fitting distributions from Table~\ref{tab:nllhkstest}.

Overall our results show that it is possible to predict RTD parameters for unseen instances, and that DistNet performed best. 
For $4$ out of $7$ datasets, our models achieved a \nllh{} close to the gold standard of fitting the RTDs to the observed data. Also, for $4$ out of $7$ datasets both distribution families were similarly easy to predict for all models.

For the RF-based models, we observed slight overfitting for most scenarios. For DistNet, we only observed this on the smallest data set: \claspfact{}.
On \spearsmall{}, both the \irf{} and \mrf{} yielded poor performance for both distributions as they failed to predict distribution parameters for instances with a very short runtime and thus receive a high \nllh{} on these instances.
In general the \irf{} yielded worse performance than \mrf{} demonstrating that distribution parameter should be learned jointly.
Overall, DistNet yielded the most robust results. It achieved the best test set predictions for all cases, sometimes with substantial improvements over the RF baselines.

\subsection{Q3: DistNet on a Low Number of Observations}

\begin{figure}[tbp]
\centering
\begin{tabular}[b]{c@{\hskip 0.1mm}c@{\hskip 0.1mm}c}
& \small{\zeno{}} & \small{\yalsathard{}}  \\
\raisebox{1\height}{\includegraphics[width=0.13\columnwidth]{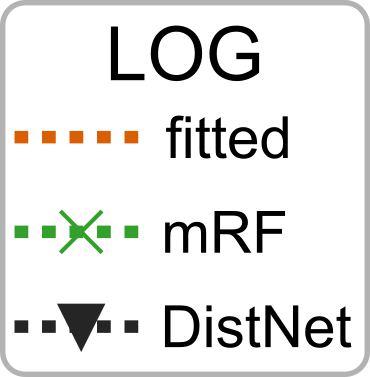}} & 
\includegraphics[width=0.35\columnwidth]{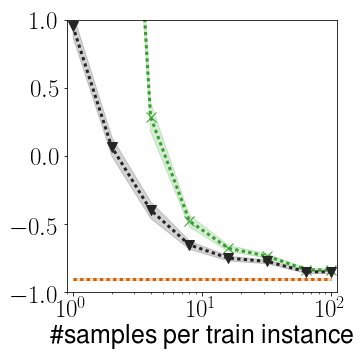} & 
\includegraphics[width=0.35\columnwidth]{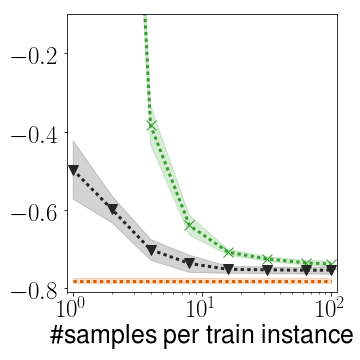} \\
\end{tabular}
\caption{\nllh{} achieved on test instances wrt. to number of observed runtimes per instance used for training. We report the mean and standard deviation across 10-folds each of which averaged across 10 repetitions. The orange line indicates the optimistic best possible \nllh{} of the fitted distribution computed on all 100 observations. \label{fig:subsets}}

\end{figure}

Finally, we evaluated the performance of DistNet wrt. the number of observed runtimes per instance. Fewer observations per instance result in smaller training data sets for DistNet, whereas the data set size for the \mrf{} stays the same with the distribution parameters being computed on fewer samples. We evaluated DistNet and \mrf{} in the same setup as before, using a $10$-fold crossvalidation, but repeating each evaluation $10$ times with a different set of sampled observations. Figure~\ref{fig:subsets} reports the achieved \nllh{} for LOG as a function of the number of training samples for two representative scenarios. We observed similar results for all scenarios.

Overall, our results show that the predictive quality of \mrf{} relies on the quality of the fitted distributions used as training data whereas DistNet can achieve better results as it directly learns from runtime observations.
On \zeno{}, for which all models performed competitively when using $100$ observations (see Table~\ref{tab:nllh}), DistNet achieved a better performance with fewer data converging to a similar \nllh{} value as the RF when using all available observations. 

On the larger dataset, \yalsathard{}, DistNet converged with $16$ samples per instance yielding a predictive performance better than that of \mrf{}s with 100 iterations. Collecting only 16 instead of 100 samples would speed up the computation by more than 6-fold, which, e.g., for \yalsathard{} would save almost 2,000 CPU hours.

\section{Conclusion and Future Work}
\label{sec:conc}

In this paper we showed that NNs can be used to jointly learn distribution parameters to predict RTDs. 
In contrast to previous RF models, we train our model on individual runtime observations, removing the need to first fit RTDs on all training instances.
More importantly, our NN -- which we dub \emph{DistNet} -- directly optimizes the loss function of interest in an end-to-end fashion, and by doing so 
obtains better predictive performance than previously-used RF models that do not directly optimize this loss function.     
Because of that our model can learn meaningful distribution parameters from only few observations per training instance and therefore requires only a fraction of training data compared to previous approaches.

Overall, our methodology allows for better RTD predictions and therefore may pave the way for improving many applications that currently rely mostly on mean predictions only, such as, e.g., algorithm selection and algorithm configuration.

Currently, our method assumes large homogeneous instance sets without censored observations. We consider further extensions as future work, such as handling censored observations (e.g., timeouts) in the loss function of our NN~\cite{gagliolo-cp06a}, using a mixture of models~\cite{jacobc-nc91a} to learn different distribution families, or studying non-parametric models (which can fit arbitrary distributions and thus do not require prior knowledge of the type of runtime distribution). Finally, in many applications the algorithm's configuration is a further important dimension for predicting RTDs, and we therefore plan to handle this as an additional input in future versions of DistNet.

\footnotesize{
{\paragraph{Acknowledgements} The authors acknowledge funding by the DFG (German Research Foundation) under Emmy Noether grant HU 1900/2-1 and support by the state of Baden-Württemberg through bwHPC and the DFG through grant no INST 39/963-1 FUGG. K.\ Eggensperger additionally acknowledges funding by the State Graduate Funding Program of Baden-Württemberg.}
}

\renewcommand{\baselinestretch}{0.92}
\bibliographystyle{named} {\footnotesize\bibliography{shortstrings,lib,local,shortproc}}

\end{document}

%% file: macros.tex
\newcommand{\notfixed}[1]{{\color{red}{#1}}}


\newcommand{\insts}[0]{\Pi}
\newcommand{\inst}[0]{\pi}
\newcommand{\instD}[0]{\mathcal{D}_\insts}
\newcommand{\pcs}[0]{\vec{\Theta}}
\newcommand{\conf}[0]{\vec{\theta}}
\newcommand{\cutoff}[0]{\kappa}
\newcommand{\loss}[0]{\mathcal{L}}

\renewcommand{\vec}[1]{\mathbf{#1}}
\newcommand{\algo}[0]{\mathcal{A}}
\newcommand{\portfolio}[0]{\mathcal{P}}
\newcommand{\feats}[0]{\mathcal{F}}
\newcommand{\feat}[0]{\vec{f}}
\newcommand{\cluster}[0]{\vec{h}}
\newcommand{\clusters}[0]{\vec{H}}
\newcommand{\perf}[0]{\mathbb{R}}
\newcommand{\surro}[0]{\hat{m}}
\newcommand{\dparam}[0]{\boldsymbol{\theta}}

\newcommand{\aclib}{\textit{AClib}} 
\newcommand{\smac}{\textit{SMAC}} 
\newcommand{\roar}{\textit{ROAR}} 
\newcommand{\paramils}{\textit{ParamILS}} 
\newcommand{\focusedils}{\textit{FocusedILS}} 
\newcommand{\pils}{\textit{PILS}}
\newcommand{\gga}{\textit{GGA}} 
\newcommand{\irace}{\textit{irace}} 
\newcommand{\spearmint}{\textit{Spearmint}} 
\newcommand{\tpe}{\textit{TPE}} 
\newcommand{\scikit}{\textit{scikit-learn}}
\newcommand{\opentuner}{\textit{OpenTuner}}
\newcommand{\nllh}{NLLH}
\newcommand{\irf}{iRF}
\newcommand{\mrf}{mRF}

\newcommand{\glucose}{\textit{Glucose}}
\newcommand{\clasp}{\textit{Clasp}}
\newcommand{\lingeling}{\textit{Lingeling}}
\newcommand{\minisathack}{\mbox{\textit{Minisat-HACK-999ED}}}
\newcommand{\minisathacksmall}{\mbox{\scriptsize\textit{Minisat-HACK-999ED}}}
\newcommand{\circuitfuzz}{\textit{Circuit Fuzz}}
\newcommand{\saps}{\textit{Saps}}
\newcommand{\cryptominisat}{\textit{Cryptominisat}}
\newcommand{\cplex}{\textit{CPLEX}}
\newcommand{\fd}{\textit{Fast Downward}} 
\newcommand{\lpg}{\textit{LPG}}
\newcommand{\spear}[0]{\textit{Spear}}
\newcommand{\minisat}[0]{\textit{MiniSATHack}}
\newcommand{\probsat}[0]{\textit{ProbSAT}}
\newcommand{\yalsat}[0]{\textit{YalSAT}}

\newcommand{\cf}[0]{\cryptominisat{}-\textit{CF}}
\newcommand{\zeno}[0]{\lpg{}-\textit{Zenotravel}}
\newcommand{\ws}[0]{\clasp{}-\textit{WS}}
\newcommand{\kfive}[0]{\clasp{}-\textit{K5}}
\newcommand{\claspfact}[0]{\clasp{}-\textit{factoring}}
\newcommand{\probsatseven}[0]{\probsat{}-\textit{7SAT}}
\newcommand{\minisatk}[0]{\minisat{}-\textit{K3}}
\newcommand{\spearswgcp}[0]{\spear{}-\textit{SWGCP}}
\newcommand{\cvfixed}[0]{\saps{}-\textit{CV-FIXED}}
\newcommand{\cvvar}[0]{\saps{}-\textit{CV-VAR}}
\newcommand{\sapsswgcp}[0]{\saps{}-\textit{SWGCP}}
\newcommand{\spearhard}[0]{\spear{}-\textit{QCP}}
\newcommand{\yalsathard}[0]{\yalsat{}-\textit{QCP}}
\newcommand{\spearsmall}[0]{\spear{}-\textit{SWGCP}}
\newcommand{\yalsatsmall}[0]{\yalsat{}-\textit{SWGCP}}
\newcommand{\spearbqcp}[0]{\spear{}-\textit{BQCP}}
\newcommand{\yalsatbqcp}[0]{\yalsat{}-\textit{BQCP}}
\newcommand{\spearback}[0]{\spear{}-\textit{QCP}}
\newcommand{\yalsatback}[0]{\yalsat{}-\textit{QCP}}
%


%% file: pipeline.tex
\tikzstyle{activity}=[rectangle, draw=black, rounded corners, text centered, text width=8em, fill=white, drop shadow]
\tikzstyle{data}=[rectangle, draw=black, text centered, fill=black!10, text width=8em, drop shadow]
\tikzstyle{myarrow}=[->, thick]
\begin{tikzpicture}[node distance=10em, scale=0.7, every node/.style={scale=0.7}]
\footnotesize
	
	\node (Algo) [data, text width=6em] {Algorithm $\algo$};
	\node (Instances) [data, text width=6em, below of=Algo, node distance=4em, text width=9em] {Training instances\\ $\inst \in \insts_{\text{train}}$};
	
	\node (Time) [activity, right of=Algo] {Run $\algo$ $k$ times on each $\inst \in \insts_{\text{train}}$};
	\node (Feats) [activity, right of=Instances] {Compute instance features $\feat(\inst)$ for each $\inst \in \insts_{\text{train}}$}; 
	
	\node (Data) [data, text width=8.1em, right of=Time, yshift=-2em] {Data:\\ $\langle \feat(\inst), t(\inst)_{\{1\ldots k\}}\rangle$};
	
	\node (EstTrain) [activity, right of=Data] {Estimate RTD family $\mathcal{D}$};
	\node (TrainFit) [activity, right of=EstTrain] {Fit RTD model\\ $\hat{m}: \feat(\inst) \mapsto \dparam$};

	\draw[myarrow] (Algo) -- (Time);
	\draw[myarrow] (Instances) -- ($(Time.west)+(0.0,-.15)$);
	\draw[myarrow] (Instances) -- (Feats);
	
	\draw[myarrow] (Feats) -- (Data);
	\draw[myarrow] (Time) -- (Data);
	
	\draw[myarrow] (Data) -- (EstTrain);
	\draw[myarrow] (EstTrain) -- (TrainFit);
	
	\draw[myarrow] (Data) |- ++(0.0,0.8) -| (TrainFit);
	
	\draw[dashed] ($(Instances)+(-2.0,-0.8)$) -- ($(Instances)+(15.5,-0.8)$);
	
	\node (NInst) [data, text width=6em, below of=Data, node distance=7em] {New instance\\ $\inst_{n+1}$};
	\node (NFeat) [activity, right of=NInst] {Compute features\\ $\feat(\inst_{n+1})$};
	\node (Predict) [activity, right of=NFeat] {Use $\hat{m}$ to predict $\mathcal{D}$'s parameters $\dparam$ for $\inst_{n+1}$};

	\draw[myarrow] (NInst) -- (NFeat);
	\draw[myarrow] (NFeat) -- (Predict);
	\draw[myarrow] (TrainFit) -- (Predict);
	
\end{tikzpicture}

%% file: ms.bbl
\begin{thebibliography}{}

\bibitem[\protect\citeauthoryear{Arbelaez \bgroup \em et al.\egroup
  }{2016}]{arbelaez-ictai16a}
A.~Arbelaez, C.~Truchet, and B.~O'Sullivan.
\newblock Learning sequential and parallel runtime distributions for randomized
  algorithms.
\newblock In {\em Proc. of ICTAI'16}, pages 655--662, 2016.

\bibitem[\protect\citeauthoryear{Babi\'c and Hutter}{2007}]{spear}
D.~Babi\'c and F.~Hutter.
\newblock Spear theorem prover, 2007.
\newblock Solver description, {SAT} competition.

\bibitem[\protect\citeauthoryear{Balint and Sch{\"o}ning}{2012}]{probSAT}
A.~Balint and U.~Sch{\"o}ning.
\newblock Choosing probability distributions for stochastic local search and
  the role of make versus break.
\newblock In {\em Proc. of SAT'12}, pages 16--29, 2012.

\bibitem[\protect\citeauthoryear{Biere}{2014}]{yalsat}
A.~Biere.
\newblock Yet another local search solver and lingeling and friends entering
  the {SAT} competition 2014.
\newblock In {\em Proc. of {SAT} Competition 2014}, pages 39--40, 2014.

\bibitem[\protect\citeauthoryear{Brewer}{1995}]{Bre95}
E.~Brewer.
\newblock High-level optimization via automated statistical modeling.
\newblock In {\em ACM SIGPLAN Notices}, pages 80--91, 1995.

\bibitem[\protect\citeauthoryear{{Chollet \textit{et al.}}}{2015}]{chollet-15a}
{Chollet \textit{et al.}}
\newblock Keras.
\newblock \url{https://github.com/fchollet/keras}, 2015.

\bibitem[\protect\citeauthoryear{Eggensperger \bgroup \em et al.\egroup
  }{2018}]{eggensperger-ml18a}
K.~Eggensperger, M.~Lindauer, H.~Hoos, F.~Hutter, and K~Leyton-Brown.
\newblock Efficient benchmarking of algorithm configuration procedures via
  model-based surrogates.
\newblock {\em Machine Learning}, 107:15--41, 2018.

\bibitem[\protect\citeauthoryear{Fawcett \bgroup \em et al.\egroup
  }{2014}]{fawcett-icaps14a}
C.~Fawcett, M.~Vallati, F.~Hutter, J.~Hoffmann, H.~Hoos, and K.~Leyton{-}Brown.
\newblock Improved features for runtime prediction of domain-independent
  planners.
\newblock In {\em Proc. of ICAPS'14}, pages 355--359, 2014.

\bibitem[\protect\citeauthoryear{Fink}{1998}]{Fink98howto}
E.~Fink.
\newblock How to solve it automatically: Selection among problem-solving
  methods.
\newblock In {\em Proc. of ICANN'08}, pages 128--136, 1998.

\bibitem[\protect\citeauthoryear{Frost \bgroup \em et al.\egroup
  }{1997}]{frost-aaai97a}
D.~Frost, I.~Rish, and L.~Vila.
\newblock Summarizing {CSP} hardness with continuous probability distributions.
\newblock In {\em Proc. of AAAI'97}, pages 327--333, 1997.

\bibitem[\protect\citeauthoryear{Gagliolo and
  Schmidhuber}{2005}]{gagliolo-icann05a}
M.~Gagliolo and J.~Schmidhuber.
\newblock A neural network model for inter-problem adaptive online time
  allocation.
\newblock In {\em Proc. of ICANN'05}, pages 752--752, 2005.

\bibitem[\protect\citeauthoryear{Gagliolo and
  Schmidhuber}{2006a}]{gagliolo-cp06a}
M.~Gagliolo and J.~Schmidhuber.
\newblock Impact of censored sampling on the performance of restart strategies.
\newblock In {\em Proc. of CP'06}, pages 167--181, 2006.

\bibitem[\protect\citeauthoryear{Gagliolo and
  Schmidhuber}{2006b}]{gagliolo-amai06a}
M.~Gagliolo and J.~Schmidhuber.
\newblock Learning dynamic algorithm portfolios.
\newblock {\em AMAI}, 47(3-4):295--328, 2006.

\bibitem[\protect\citeauthoryear{Gebser \bgroup \em et al.\egroup
  }{2012}]{gebser-ai12}
M.~Gebser, B.~Kaufmann, and T.~Schaub.
\newblock Conflict-driven answer set solving: From theory to practice.
\newblock {\em AI}, 187-188:52--89, 2012.

\bibitem[\protect\citeauthoryear{Gent \bgroup \em et al.\egroup
  }{1999}]{gent-aaai99}
I.~Gent, H.~Hoos, P.~Prosser, and T.~Walsh.
\newblock Morphing: Combining structure and randomness.
\newblock In {\em Proc. of AAAI'99}, pages 654--660, 1999.

\bibitem[\protect\citeauthoryear{Gerevini and Serina}{2002}]{gerevini-aips02}
A.~Gerevini and I.~Serina.
\newblock {LPG:} {A} planner based on local search for planning graphs with
  action costs.
\newblock In {\em Proc. of AIPS'02}, pages 13--22, 2002.

\bibitem[\protect\citeauthoryear{Gomes and Selman}{1997}]{gomes-aaai97}
C.~Gomes and B.~Selman.
\newblock Problem structure in the presence of perturbations.
\newblock In {\em Proc. of AAAI'97}, pages 221--226, 1997.

\bibitem[\protect\citeauthoryear{Gomes and Selman}{2001}]{gomes-aij01}
C.~Gomes and B.~Selman.
\newblock Algorithm portfolios.
\newblock {\em AIJ}, 126(1-2):43--62, 2001.

\bibitem[\protect\citeauthoryear{Gomes \bgroup \em et al.\egroup
  }{2000}]{gomes-jar00a}
C.~Gomes, B.~Selman, N.~Crato, and H.~Kautz.
\newblock Heavy-tailed phenomena in satisfiability and constraint satisfaction
  problems.
\newblock {\em JAR}, 24:67--100, 2000.

\bibitem[\protect\citeauthoryear{Goodfellow \bgroup \em et al.\egroup
  }{2016}]{goodfellow-mit16a}
I.~Goodfellow, Y.~Bengio, and A.~Courville.
\newblock {\em Deep Learning}.
\newblock MIT Press, 2016.

\bibitem[\protect\citeauthoryear{Haim and Walsh}{2009}]{haim-sat09a}
S.~Haim and T.~Walsh.
\newblock Restart strategy selection using machine learning techniques.
\newblock In {\em Proc. of SAT'09}, pages 312--325, 2009.

\bibitem[\protect\citeauthoryear{Hoos and Stützle}{2004}]{hoos-sls04}
H.~Hoos and T.~Stützle.
\newblock {\em Stochastic Local Search: Foundations \& Applications}.
\newblock Morgan Kaufmann Publishers Inc., 2004.

\bibitem[\protect\citeauthoryear{Hurley and O'Sullivan}{2015}]{hurley-ijcai15a}
B.~Hurley and B.~O'Sullivan.
\newblock Statistical regimes and runtime prediction.
\newblock In {\em Proc. of IJCAI'15}, pages 318--324, 2015.

\bibitem[\protect\citeauthoryear{Hutter \bgroup \em et al.\egroup
  }{2002}]{saps}
F.~Hutter, D.~Tompkins, and H.~Hoos.
\newblock Scaling and probabilistic smoothing: Efficient dynamic local search
  for {SAT}.
\newblock In {\em Proc.of CP'02}, pages 233--248, 2002.

\bibitem[\protect\citeauthoryear{Hutter \bgroup \em et al.\egroup
  }{2006}]{hutter-cp06a}
F.~Hutter, Y.~Hamadi, H.~Hoos, and K.~Leyton-Brown.
\newblock Performance prediction and automated tuning of randomized and
  parametric algorithms.
\newblock In {\em Proc. of CP'06}, pages 213--228, 2006.

\bibitem[\protect\citeauthoryear{Hutter \bgroup \em et al.\egroup
  }{2011}]{hutter-lion11a}
F.~Hutter, H.~Hoos, and K.~Leyton-Brown.
\newblock Sequential model-based optimization for general algorithm
  configuration.
\newblock In {\em Proc. of LION'11}, pages 507--523, 2011.

\bibitem[\protect\citeauthoryear{Hutter \bgroup \em et al.\egroup
  }{2013}]{hutter-lion13a}
F.~Hutter, H.~Hoos, and K.~Leyton-Brown.
\newblock Identifying key algorithm parameters and instance features using
  forward selection.
\newblock In {\em Proc. of LION'13}, pages 364--381, 2013.

\bibitem[\protect\citeauthoryear{Hutter \bgroup \em et al.\egroup
  }{2014}]{hutter-aij14a}
F.~Hutter, L.~Xu, H.~Hoos, and K.~Leyton-Brown.
\newblock Algorithm runtime prediction: Methods and evaluation.
\newblock {\em AIJ}, 206:79--111, 2014.

\bibitem[\protect\citeauthoryear{Jacobs \bgroup \em et al.\egroup
  }{1991}]{jacobc-nc91a}
R.~Jacobs, M.~Jordan, S.~Nowlan, and G.~Hinton.
\newblock Adaptive mixtures of local experts.
\newblock {\em Neural Computation}, 3(1):79--87, 1991.

\bibitem[\protect\citeauthoryear{Jones \bgroup \em et al.\egroup
  }{2001}]{jones-01a}
E.~Jones, T.~Oliphant, and P.~Peterson \textit{et al.}
\newblock {SciPy}: Open source scientific tools for {Python}.
\newblock \url{http://www.scipy.org/}, 2001.

\bibitem[\protect\citeauthoryear{Leyton-Brown \bgroup \em et al.\egroup
  }{2009}]{leyton-brown-acm09a}
K.~Leyton-Brown, E.~Nudelman, and Y.~Shoham.
\newblock Empirical hardness models: Methodology and a case study on
  combinatorial auctions.
\newblock {\em Journal of ACM}, 56(4):1--52, 2009.

\bibitem[\protect\citeauthoryear{Loreggia \bgroup \em et al.\egroup
  }{2016}]{loreggia-aaai16}
A.~Loreggia, Y.~Malitsky, H.~Samulowitz, and V.~Saraswat.
\newblock Deep learning for algorithm portfolios.
\newblock In {\em Proc. of AAAI'16}, pages 1280--1286, 2016.

\bibitem[\protect\citeauthoryear{Luby \bgroup \em et al.\egroup
  }{1993}]{luby-ipl93}
M.~Luby, A.~Sinclair, and D.~Zuckerman.
\newblock Optimal speedup of las vegas algorithms.
\newblock {\em Inf. Process. Lett.}, pages 173--180, 1993.

\bibitem[\protect\citeauthoryear{Nudelman \bgroup \em et al.\egroup
  }{2004}]{nudelman-cp04}
E.~Nudelman, K.~Leyton-Brown, A.~Devkar, Y.~Shoham, and H.~Hoos.
\newblock Understanding random {SAT}: Beyond the clauses-to-variables ratio.
\newblock In {\em Proc. of CP'04}, pages 438--452, 2004.

\bibitem[\protect\citeauthoryear{Pascanu \bgroup \em et al.\egroup
  }{2014}]{pascanu-iclm13a}
R.~Pascanu, T.~Mikolov, and Y.~Bengio.
\newblock On the difficulty of training recurrent neural networks.
\newblock In {\em Proc. of ICML'13}, pages 1310--1318, 2014.

\bibitem[\protect\citeauthoryear{Pedregosa \bgroup \em et al.\egroup
  }{2011}]{scikit-learn}
F.~Pedregosa, G.~Varoquaux, A.~Gramfort, V.~Michel, B.~Thirion, O.~Grisel,
  M.~Blondel, P.~Prettenhofer, R.~Weiss, V.~Dubourg, J.~Vanderplas, A.~Passos,
  D.~Cournapeau, M.~Brucher, M.~Perrot, and E.~Duchesnay.
\newblock Scikit-learn: Machine learning in {P}ython.
\newblock {\em JMLR}, 12:2825--2830, 2011.

\bibitem[\protect\citeauthoryear{Penberthy and Weld}{1994}]{penberthy-aaai94}
J.~Penberthy and D.~Weld.
\newblock Temporal planning with continuous change.
\newblock In {\em Proc. of AAAI'94}, pages 1010--1015, 1994.

\bibitem[\protect\citeauthoryear{Roberts and
  Howe}{2007}]{Roberts07learnedmodels}
Mark Roberts and Adele Howe.
\newblock Learned models of performance for many planners.
\newblock In {\em ICAPS 2007 Workshop AI Planning and Learning}, 2007.

\bibitem[\protect\citeauthoryear{Smith-Miles and Lopes}{2012}]{SmiLop12}
K.~Smith-Miles and L.~Lopes.
\newblock Measuring instance difficulty for combinatorial optimization
  problems.
\newblock {\em Computers and Operations Research}, 39(5):875--889, 2012.

\bibitem[\protect\citeauthoryear{Xu \bgroup \em et al.\egroup
  }{2008}]{xu-jair08a}
L.~Xu, F.~Hutter, H.~Hoos, and K.~Leyton-Brown.
\newblock {SAT}zilla: Portfolio-based algorithm selection for {SAT}.
\newblock {\em JAIR}, 32:565--606, 2008.

\end{thebibliography}
